\documentclass[letterpaper, 10 pt, conference]{ieeeconf}  

\usepackage{times}
\usepackage{epsfig}
\usepackage{graphicx}
\usepackage{amsmath}
\usepackage{amssymb}
\usepackage{algorithmic}
\usepackage{multirow}
\usepackage{xcolor}
\usepackage{soul}
\usepackage{url}
\usepackage[linesnumbered,boxed,ruled,commentsnumbered]{algorithm2e}

\newtheorem{theorem}{Theorem}[section]
\newtheorem{prop}{Proposition}

\newtheorem{lemma}[theorem]{Lemma}

\IEEEoverridecommandlockouts                              

\title{\LARGE \bf
Collaborative Multi-Object Tracking \\with Conformal Uncertainty Propagation
}

\author{Sanbao Su$^{1}$, Songyang Han$^{3}$, Yiming Li$^{2}$, Zhili Zhang$^{1}$, Chen Feng$^{2}$, Caiwen Ding$^{1}$, and Fei Miao$^{1}$
\thanks{Sanbao Su, Songyang Han, Zhili Zhang and Fei Miao are supported by the National Science Foundation under Grants CNS-1952096 and CNS-2047354. Chen Feng is supported by the National Science Foundation under Grants FRR-2238968, and CPS-2121391.
$^{1}$Sanbao Su, Zhili Zhang, Caiwen Ding, and Fei Miao are with the Department of Computer Science and Engineering, University of Connecticut, Storrs Mansfield, CT, USA 06268 {\tt\small\{sanbao.su, zhili.zhang, caiwen.ding, fei.miao\}@uconn.edu}. $^{2}$Yiming Li and Chen Feng are with Tandon School of Engineering, New York University, Brooklyn, NY, USA 11201 {\tt\small\{yimingli, cfeng\}@nyu.edu}. 
$^{3}$Songyang Han is now with Sony AI, {\tt\small songyang.han@sony.com} and this work was done when he was a PhD candidate at the University of Connecticut.
}
}

\begin{document}

\maketitle
\thispagestyle{empty}
\pagestyle{empty}

\begin{abstract}
   Object detection and multiple object tracking (MOT) are essential components of self-driving systems. Accurate detection and uncertainty quantification are both critical for onboard modules, such as perception, prediction, and planning, to improve the safety and robustness of autonomous vehicles. Collaborative object detection (COD) has been proposed to improve detection accuracy and reduce uncertainty by leveraging the viewpoints of multiple agents. However, little attention has been paid to how to leverage the uncertainty quantification from COD to enhance MOT performance. In this paper, as the first attempt to address this challenge, we design an uncertainty propagation framework called MOT-CUP. Our framework first quantifies the uncertainty of COD through direct modeling and conformal prediction, and propagates this uncertainty information into the motion prediction and association steps. MOT-CUP is designed to work with different collaborative object detectors and baseline MOT algorithms. We evaluate MOT-CUP on V2X-Sim,  a comprehensive collaborative perception dataset, and demonstrate a $2\%$ improvement in accuracy and a $2.67\times$ reduction in uncertainty compared to the baselines, e.g. SORT and ByteTrack. In scenarios characterized by high occlusion levels, our MOT-CUP demonstrates a noteworthy $4.01\%$ improvement in accuracy. MOT-CUP demonstrates the importance of uncertainty quantification in both COD and MOT, and provides the first attempt to improve the accuracy and reduce the uncertainty in MOT based on COD through uncertainty propagation. Our code is public on \url{https://coperception.github.io/MOT-CUP/}.
\end{abstract}

\section{Introduction}
\label{sec:intro}

 Object detection~\cite{feng2021review} and multiple object tracking (MOT)~\cite{Luo_2021_ICCV} represent crucial steps of self-driving, and their accuracy and uncertainty quantification (UQ) are important to facilitate various onboard modules including perception, prediction and planning, to improve the safety and robustness of the autonomous systems~\cite{boris2022propagating, han2022behavior, mpUncertain_icra14}. Multi-agent collaborative object detection (COD) has been proposed to leverage the viewpoints of multiple agents to enhance detection accuracy compared with individual viewpoints~\cite{li2022v2x,xu2022opencood}. Numerous studies have demonstrated the advantages of COD in enhancing the detection accuracy~\cite{arnold2020cooperative,Chen2019CooperCP,li2021learning,li2022multi} and reducing the uncertainty~\cite{Su2022uncertainty}. Currently, Tracking-by-Detection is considered as one of the most effective paradigms~\cite{zhang2021bytetrack}, using Kalman Filter to predict the next location based on the previous detection results and then performing data association~\cite{bewley2016simple, cao2022observation, pang2021quasi, wojke2017simple}.

\begin{figure}
    \centering
    \includegraphics[width=1\linewidth]{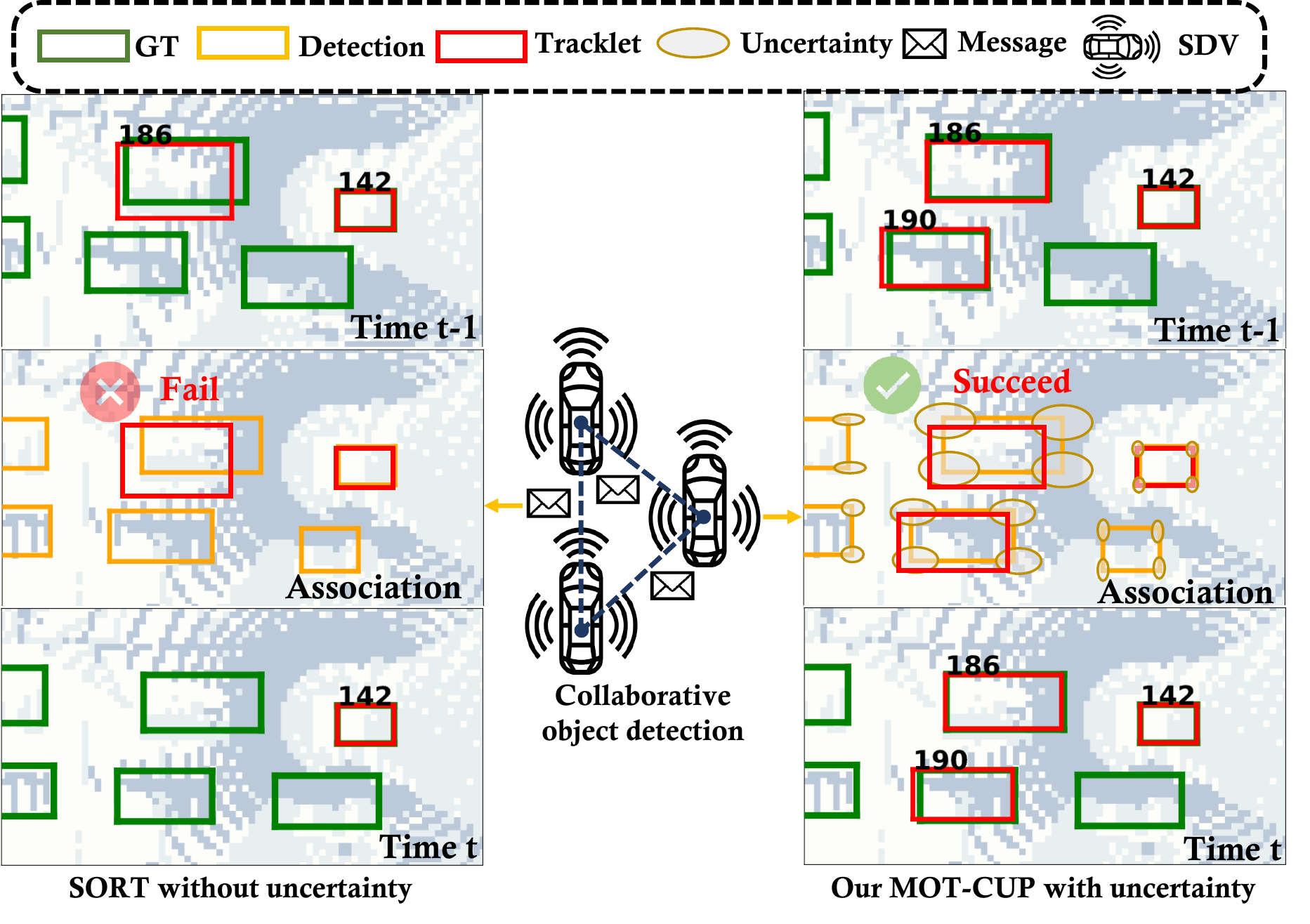}
    \vspace{-8mm}
    \caption{\textbf{Difference in data association for MOT with and without considering uncertainty}. Ground truth bounding boxes are in green, detected bounding boxes in orange, and tracklets' bounding boxes in red, labeled with object IDs. Shadow ellipses indicate uncertainty of the detected bounding box. SORT~\cite{bewley2016simple}, which doesn't consider uncertainty, is on the left side of the figure, while our MOT-CUP framework, which incorporates uncertainty, is on the right side. At time (t-1), both MOT algorithms output tracklet ID 186. However, at time t, SORT fails to associate the low-quality detection with tracklet 186 due to a large IoU distance. Thus, SORT removes the tracklet. In contrast, our MOT-CUP framework quantifies the uncertainty of COD with a larger shadow ellipse to represent the uncertainty of the bounding box for tracklet 186, and successfully associates the low-quality detection by considering the uncertainty of COD.}
    \label{fig:teaser}
    \vspace{-8mm}
\end{figure}

However, limitations exist in the methods mentioned above. Existing Kalman Filter (KF) algorithms for motion prediction typically use a fixed measured uncertainty for all detections instead of rigorously calculated uncertainty. Moreover, employing the Intersection-over-Union (IoU) association metric without considering uncertainty in the Hungarian algorithm might not suit poorly detected results due to occlusion. Hence, it remains challenging to rigorously quantify and propagate the uncertainty from COD to MOT to improve the accuracy, for both KF and association steps. For instance, Fig.~\ref{fig:teaser} illustrates how our framework outperforms SORT~\cite{bewley2016simple} in associating tracklet 186 (red box) with the low-quality object detection (orange box) at time $t$. The IoU metric in SORT fails to match them due to the poor detection quality; whereas our framework, by incorporating detection uncertainty (shadow ellipses), effectively associates tracklet 186 even with the low-quality detection. It demonstrates that integrating uncertainty into MOT can improve tracking performance, especially for low-quality detection scenarios.

In this paper, we propose a novel uncertainty propagation framework to improve the performance of these Tracking-by-Detection algorithms, called MOT-CUP (Multiply Object Tracking with Conformal Uncertainty Propagation). Specifically, our proposed MOT-CUP involves uncertainty quantification of collaborative object detection (COD) via direct modeling and conformal prediction techniques. 
The uncertainty obtained from the COD stage is subsequently incorporated into both the Kalman Filter and the association procedure of MOT. In particular, we define a new association metric with Negative Log Likelihood (NLL) considering the uncertainty of COD and potential low-quality detection results. Through extensive experiments on V2X-SIM~\cite{li2022v2x} and a series of Tracking-by-Detection MOT algorithms, we show that MOT-CUP framework improves accuracy with up to 2\% and reduces uncertainty with up to $2.67\times$. In high occlusion-level scenarios, our MOT-CUP achieves a $4.01\%$ improvement in accuracy. This outcome underscores the effectiveness of our MOT-CUP in challenging scenarios with poor detection. Our results also provide strong validation for the effectiveness of rigorous conformal prediction-based uncertainty quantification in MOT. Overall, our findings highlight the potential benefits of propagating uncertainty quantification into MOT algorithms.

The main contributions of this work are as follows:
\begin{enumerate}
    \item To the best of our knowledge, our MOT-CUP framework is the first attempt to leverage quantified uncertainty from collaborative object detection to improve MOT performance. This framework can be applied to most object detection models and MOT algorithms.
    \item In the collaborative object detection stage, we employ direct modeling and conformal prediction techniques to rigorously quantify the uncertainty.
    \item For MOT, we further improve the original MOT algorithm by designing two novel methods that effectively leverage uncertainty information for both the Kalman Filter and association.
\end{enumerate}


\section{Related Work}
\label{sec:relatedwork}




\paragraph{Collaborative Object Detection (COD)}\label{subsec:cod}
Collaborative Object Detection (COD) surpasses traditional Single-Agent Object Detection (SOD) by harnessing information from multiple agents or sensors, elevating detection accuracy \cite{li2022v2x, xu2022opencood, cai2022analyzing} and mitigating uncertainty \cite{Su2022uncertainty}. Multi-camera object detection (MOD) \cite{jiang2023polarformer} methods use strategically positioned cameras of a single agent to improve performance~\cite{jiang2023polarformer}. In complex scenarios such as low-light conditions, occlusions, and adverse weather, COD outperforms MOD for sharing complementary information by communication between multiple agents~\cite{hu2022where2comm}. Its dynamic integration of insights from various sources enables effective adaptation to changing environments and scene dynamics. Hence, COD harmonizes viewpoints and enables innovative occlusion handling. Moreover, COD extends coverage and precision, with well-designed orchestration to avert redundancy or misalignment~\cite{li2022v2x}. In summary, COD redefines object detection, leveraging collaboration to tackle challenges and chart the future of detection~\cite{xu2022v2x}. 
In this study, we want to show that even though object detection performance has already improved, uncertainty propagation from an advanced model such as COD is still important to enhance the overall performance of subsequent modules.

\paragraph{Uncertainty Quantification and Propagation}\label{subsec:uq}

Uncertainty Quantification (UQ) is vital for the collaborative perception of safety-critical systems such as robots~\cite{rss19, perceptionCBF_corl21} and connected and autonomous vehicles~\cite{han2022solution, han2022stable, he2023robust}. 
In self-driving tasks, UQ from COD could improve trajectory prediction~\cite{boris2022propagating} and motion planning~\cite{mpUncertain_icra14}. However, there is no research on how to leverage COD uncertainty to enhance tracking performance. Several methods for UQ in object detection (OD) require multiple inference runs, not designed for real-time tasks like COD, such as Monte-Carlo dropout~\cite{miller2018dropout} and deep ensemble~\cite{lakshminarayanan2017simple,lyu2020probabilistic}. 


Direct modeling (DM)~\cite{feng2021review} methods for OD have been proposed~\cite{Su2022uncertainty, meyer2020learning}. DM is promising for real-time computer vision tasks, as only requires a single inference pass. However, DM lacks rigorous UQ as the model may easily overfit the training dataset. The work in~\cite{Su2022uncertainty} proposes the bootstrap calibrated DM method for COD. However, the bounding box definition it presented, which relies on corner coordinates, is not congruent with tracking algorithms. Conformal prediction (CP)~\cite{angelopoulos2021gentle} is a statistical method that converts any heuristic notion of uncertainty (e.g. standard deviation estimations) into rigorous UQ. To rigorously quantify the uncertainty in COD and propagate the uncertainty to MOT, our MOT-CUP framework leverages CP to calibrate the uncertainty estimation from DM.



\paragraph{Multiple Object Tracking (MOT)}\label{subsec:mot}

Several recent MOT algorithms~\cite{zhang2021bytetrack,cao2022observation,wojke2017simple,choi2015near,zhou2020tracking} use motion models based on Bayesian estimation~\cite{lehmann2006theory} to predict states by maximizing the posterior estimation. Kalman Filter (KF)~\cite{kalman1960contributions}, a widely used motion model, operates as a recursive Bayes filter that follows a standard predict-update cycle. 
Current KF-based algorithms typically use a fixed measured uncertainty for all detections without considering rigorously estimated uncertainty to improve the prediction accuracy. In contrast, we propose a rigorous UQ of the COD process based on CP, and integrating it into KF for enhanced accuracy and uncertainty estimation of output.

Data association is a crucial step of MOT, which involves computing the similarity between tracklets and detected objects and utilizing various strategies to match them based on their similarity. 
The SORT algorithm~\cite{bewley2016simple} uses the Intersection over Union (IoU) between predicted and detected boxes to determine their similarity. This approach has proven to be highly competitive on a variety of MOT benchmarks, and serves as a strong baseline for more sophisticated tracking methods.
With the similarity, matching strategies assign identities to the objects. This can be achieved through the Hungarian Algorithm~\cite{kuhn1955hungarian} or greedy assignment~\cite{zhou2020tracking}. 
ByteTrack~\cite{zhang2021bytetrack} utilizes similarities between the low-quality detections and tracklets to recover accurate object identities,  improving data association performance.
However, using IoU distance as a similarity metric may not be appropriate for low-quality detections as explained in Fig.~\ref{fig:teaser}. 
The Mahalanobis distance is also a widely used similarity score by quantifying the dissimilarity between the detected objects and the distributions of tracklets from the KF model~\cite{bertozzi2004pedestrian}. Nonetheless, substantial uncertainty could result in minimal distances, leading to erroneous matching~\cite{altendorfer2016association,wojke2017simple}. 
The work in \cite{altendorfer2016association} proposes the association log-likelihood distance to overcome this problem by computing the logarithmized association probability.
Unlike conventional similarity scores, we propose the Negative Log Likelihood (NLL) similarity score, which computes the NLL between the detected distribution of the objects and the mean of the tracklets to focus on the distribution of detections. So it can ignore the large uncertainty and facilitate more accurate associations for low-quality detections.


\section{Methodology}
\label{sec:approach}

\subsection{Approach Overview} 

\begin{figure}
    \centering
    \includegraphics[width=1\linewidth]{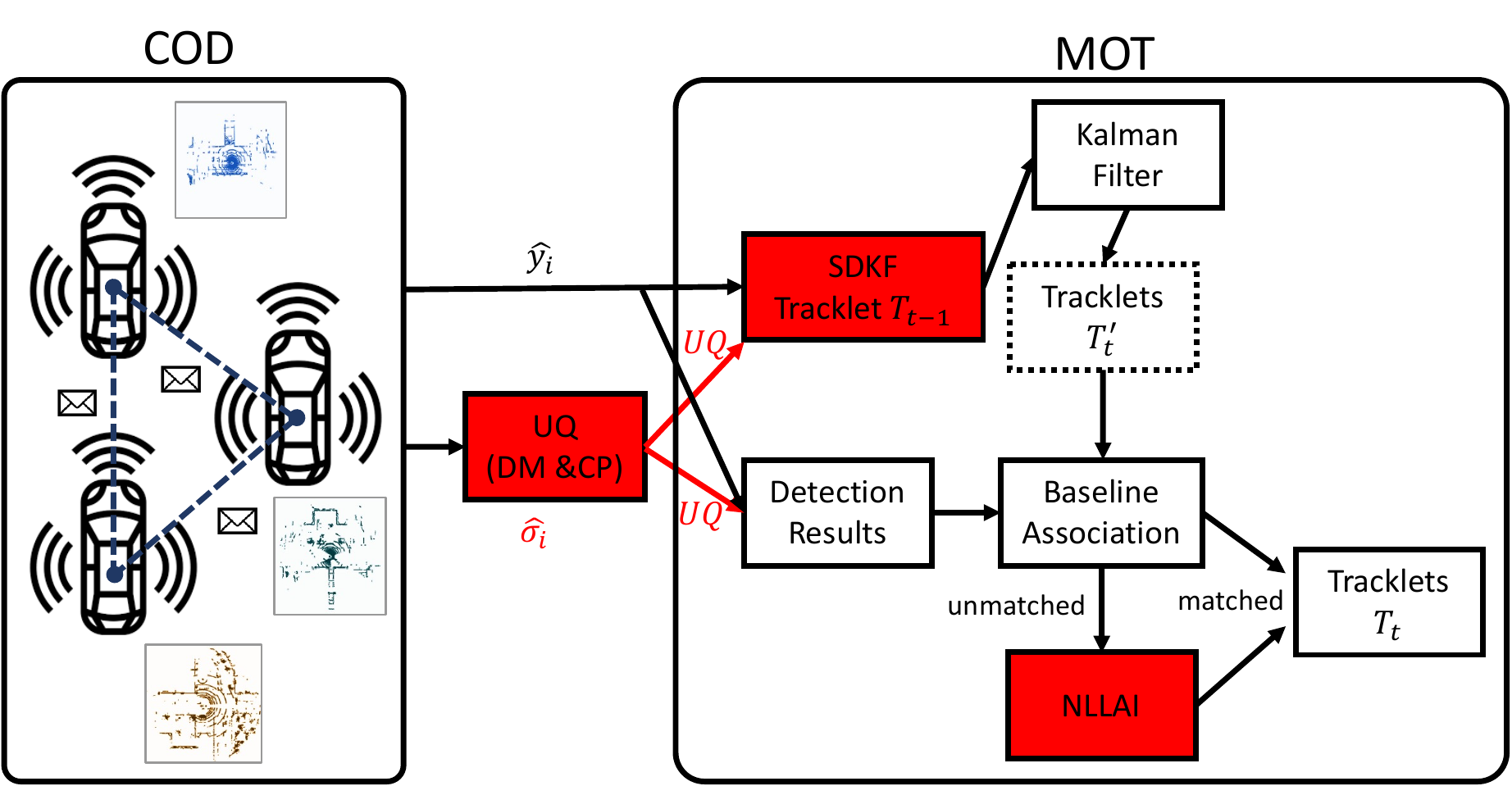}
    \caption{\textbf{Overview of our MOT-CUP framework.} The red color highlights the novelties and important techniques in our MOT-CUP framework. In the collaborative object detection (COD) stage, we rigorously calculate uncertainty quantification (UQ) of each object detection via direct modeling (DM) and conformal prediction (CP).  In the motion prediction stage of MOT, we adopt a Standard Deviation-based Kalman Filter (SDKF) to enhance the Kalman Filter process, that leverages the UQ results and predicts the locations of the objects in the next time step with higher precision. In the association step, we first apply the baseline association method and then associate the unmatched detections and tracklets with the Negative Log Likelihood similarity metric, called NLLAI.}
    \label{fig:overview}
    \vspace{-5mm}
\end{figure}

We design a novel framework for uncertainty propagation of collaborative object detection (COD) to MOT, named Multiple Object Tracking with Conformal Uncertainty Propagation (MOT-CUP). Fig.~\ref{fig:overview} presents the methodology overview. The major novelties are: \textit{(1)} MOT-CUP framework first rigorously quantifies the uncertainty in the COD stage based on direct modeling and conformal prediction. \textit{(2)} The uncertainty information is leveraged in the motion prediction stage of MOT, where a Standard Deviation-based Kalman Filter (SDKF) takes the uncertainty quantification (UQ) of COD as its input to improve the predicted precision of location. \textit{(3)} We utilize the Negative Log Likelihood as the similarity metric for the association step, called NLLAI, to improve the accuracy and reduce the uncertainty of MOT.

In this section, we first introduce the conformal prediction in Subsection~\ref{subsec:pre} as preliminary literature of UQ and a useful method to construct predicted uncertainty. We describe our proposed MOT-CUP (Multiply Object Tracking with Conformal Uncertainty Propagation) method as shown in Algorithm~\ref{alg:whole} and Subsection~\ref{subsec:our_app}, followed by the detailed process of UQ of COD based on direct modeling and conformal prediction in Subsection~\ref{subsec:uq_detection}, and uncertainty propagation to MOT in Subsection~\ref{subsec:uq_track}. 

\subsection{Preliminary}\label{subsec:pre} 
Conformal prediction (CP)~\cite{angelopoulos2021gentle} is a statistical method to generate prediction sets for any model. It is a method to convert any heuristic notion of uncertainty (e.g. an estimate of the standard deviation) to rigorous UQ. 
For example, we assume that an uncertain scalar follows Gaussian distribution and train a model to output the mean and standard deviation. To be precise, we choose to model $Y_{test} \sim \mathcal{N}(\mu(x), \sigma(x)) \mid_{X_{test} = x}$, where $X_{test}$ is a testing data, and $Y_{test}$ is the corresponding label. We train $\hat{\mu}(x)$ and $\hat{\sigma}(x)$ to maximize the likelihood of the data. Conformal prediction can turn this heuristic uncertainty notion into rigorous prediction intervals of the form $(\hat{\mu}(x) \pm \hat{q}\hat{\sigma}(x))$, where $\hat{q}$ is a quantile found by CP. 

Consider the validation data $(X_1, Y_1), ..., (X_N, Y_N)$ with $N$ data points that are never seen during training, the CP for input $x$ and output $y$ includes the following steps:
\textit{(1)} Define the score function $s(x, y) \in \mathbb{R}$. (Smaller scores encode better agreement between $x$ and $y$). \textit{(2)} Compute $\hat{q}$ as the $\frac{\lceil (N+ 1) (1-\alpha) \rceil}{N}$ quantile of the validation scores $s_1 = s(X_1, Y_1), ..., s_N = s(X_N, Y_N)$, where $\alpha \in [0, 1]$ is a user-chosen error rate. \textit{(3)} Use this quantile to form the prediction sets $\mathcal{C}(X_{test})$ for new examples:
\vspace{-5pt}
\begin{equation}
    \label{equ:prediction_set}
        \mathcal{C}(X_{test}) = \{ y:s(X_{test}, y) \leq \hat{q}\},
\end{equation}
Note that $(X_{test}, Y_{test})$ is a fresh test point from the same distributions of the validation data. The CP provides a coverage guarantee, as stated in the following lemma.

\begin{lemma}[\textbf{Conformal Coverage Guarantee}~\cite{angelopoulos2021gentle}]
\label{lemma:coverage}
Suppose $(X_k, Y_k)_{k=1, ..., N}$ and $(X_{test}, Y_{test})$ are $i.i.d.$, then the following holds:
\vspace{-5pt}
\begin{equation}
    1- \alpha \leq \Pr(Y_{test} \in \mathcal{C}(X_{test})) \leq 1- \alpha + \frac{1}{N + 1}.
\end{equation}
\end{lemma}

In other words, the probability that the prediction set contains the correct label is almost exactly $1-\alpha$. 

\subsection{MOT-CUP Algorithm} 
\label{subsec:our_app}


The detail of MOT-CUP is presented in Algorithm~\ref{alg:whole}. 
For each frame in the point cloud sequence $S$, there are $J$ objects. For each frame, the trained collaborative object detector with direct modeling would generate a set of detected objects $\mathcal{D} = \{ \hat{p}_j,  loc_j \}_{j=1}^J$ (Line~\ref{alg:line_detector}). The set includes the predicted classification probability $\hat{p}_j$ and the location of each object $loc_j$. The location of each object is represented by $I$ random variables parameterized by $\{\hat{y}_{i}, \hat{\sigma}_{i}\}_{i=1}^I$ where  $\hat{y}_{i}$ is the mean and $\hat{\sigma}_{i}$ is the standard deviation for $i$-th variable. This object detector not only predicts the location of each object but also provides a measure of uncertainty.

To provide more accurate measures of uncertainty, we leverage the quantiles computed by CP to adjust the standard deviation $\hat{\sigma}_{i}$ (see Lines~\ref{alg:line_cp_begin}-\ref{alg:line_cp_end}). Then, to track the detected objects across multiple frames, we employ a Kalman Filter to predict the current state of the tracklets, which is commonly used in MOT~\cite{bewley2016simple} (see Lines~\ref{alg:line_pred_tracklet_begin}-\ref{alg:line_pred_tracklet_end}). In the association step, we first apply the origin association method (see Line~\ref{alg:line_ba}) and store all matched pairs $(d, t)$ in $\mathcal{A}_{matched}$ for $d \in \mathcal{D}$ and $t\in \mathcal{T}$. Then we associate the unmatched detections and tracklets with the Negative Log Likelihood similarity metric for lower-quality detected objects (see Line~\ref{alg:line_nllai}). The detail of Negative Log Likelihood-based Association Improvement (NLLAI) will be introduced in Algorithm~\ref{alg:nllai}. To update the tracklets with the matched detections, we go beyond the traditional MOT algorithms by incorporating the detected standard deviation $\hat{\sigma}_i$ in addition to the detected mean $\hat{y}_i$ (see Lines~\ref{alg:line_update_begin}-\ref{alg:line_update_end}). This allows us to more accurately model the uncertainty associated with each detection and incorporate it into the tracklet. The detected standard deviations are also applied to generating new tracklets with unmatched detections (see Lines~\ref{alg:line_add_tracklet_begin}-\ref{alg:line_add_tracklet_end}). 
\vspace{-10pt}
\begin{algorithm}[ht]
\small
\caption{Multiple Object Tracking with Conformal Uncertainty Propagation (MOT-CUP)}
\label{alg:whole}
\KwData{input point cloud sequence $S$, the trained collaborative object detector $F$, NLL threshold $\tau$, baseline's Kalman Filter $KF$, baseline's association method $BA$, quantile of CP $\hat{q}$}
\KwResult{Tracklet list $\mathcal{T}$}
Initialization: $\mathcal{T} \leftarrow \emptyset$ \\
\For{point cloud frame $C$ in $S$ in time sequence}{
    $\mathcal{D} = F(C) = \{ \hat{p}_j,  loc_j \}_{j=1}^J$, where each location contains $\{\hat{y}_{i}, \hat{\sigma}_{i}\}_{i=1}^I$ \label{alg:line_detector}\\
    \For{each object}{ \label{alg:line_cp_begin}
    \For{i\ from\ 1\ to\ I}{
    $\hat{\sigma}_{i} = \hat{\sigma}_{i} \times \hat{q}_i$
    }
    }\label{alg:line_cp_end}
    \tcc{Adjust standard deviation by \textbf{CP}}
    \For{$t$\ in\ $\mathcal{T}$}{\label{alg:line_pred_tracklet_begin}
    Apply Kalman Filter (KF)~\cite{kalman1960contributions}\\
    }\label{alg:line_pred_tracklet_end}
    $\mathcal{A}_{matched}, \mathcal{D}_{unmatched}, \mathcal{T}_{unmatched} = BA(\mathcal{D}, \mathcal{T})$\label{alg:line_ba}\\
    $\mathcal{A}'_{matched}, \mathcal{D}'_{unmatched}, \mathcal{T}'_{unmatched} = \text{NLLAI}(\mathcal{A}_{matched}, \mathcal{D}_{unmatched}, \mathcal{T}_{unmatched}, \tau)$\label{alg:line_nllai}\\
    \tcc{\textbf{NLLAI} is Algorithm~\ref{alg:nllai}}
    \For{$(d, t)$ in $\mathcal{A}'_{matched}$}{\label{alg:line_update_begin}
        Apply KF with updated standard deviation $\hat{\sigma}$\\
    }\label{alg:line_update_end}
    $\mathcal{T} = \mathcal{T} \backslash \mathcal{T}'_{unmatched}$\\
    \For{$d$ in $\mathcal{D}'_{unmatched}$}{\label{alg:line_add_tracklet_begin}
    $\mathcal{T} = \mathcal{T} \cup \{d\}$ where $d=\{\hat{y}_{i},\hat{\sigma}_{i}\}_{i=1}^I$\\
    }\label{alg:line_add_tracklet_end}
}
\end{algorithm}\vspace{-15pt}

\subsection{UQ on Collaborative Object Detection}\label{subsec:uq_detection}
We use direct modeling~\cite{Su2022uncertainty,he2019bounding} to estimate the standard deviation of each variable of the COD stage. We assume that all variables are independent and the distribution of each variable is a single-variate Gaussian distribution. For the distribution of each variable of the ground truth, we assume it as a Dirac delta function~\cite{he2019bounding}. Then we define the regression loss function for the $i$-th variable as the Kullback-Leibler (KL) divergence between the single-variate Gaussian distribution and the Dirac delta function~\cite{murphy2012machine}:

\vspace{-10pt}
\begin{equation}
    \mathcal{L}^i_{KL}(y_i, \hat{y}_i, \hat{\sigma}_i) =  \frac{(y_i-\hat{y}_i)^2}{2\hat{\sigma}_i^2}  + \log |\hat{\sigma}_i|,
\label{eq:kl-loss}
\end{equation}
where $y_i$ is the ground-truth value for $i$-th variable. An additional regression header is incorporated to forecast all standard deviations $\hat{\sigma}_i$, with a comparable structure as the regression header for $\hat{y}_i$. This is accomplished based on the original collaborative object detector where no alterations have been made to the remaining components. 

After we have the trained object detection model, we compute the quantile for the standard deviation of each variable by CP~\cite{angelopoulos2021gentle} based on the validation dataset, as introduced in Subsection~\ref{subsec:pre}.

We define the score function for the $i$-th variable as:
\begin{equation}
    s(x_i, y_i) = \frac{|y_i - \hat{y}_i(x_i)|}{\hat{\sigma}_i(x_i)},
\label{eq:score_func}
\end{equation}
where $x_i$ is the point cloud input and it can be comprehended as a multiplicative correction factor applied to the standard deviation where $s(x_i, y_i) \hat{\sigma}_i(x_i) = |y_i - \hat{y}_i(x_i)|$. After testing the detection model on the validation dataset and calculating the score function, we obtained a set of scores $\{s_{i1},s_{i2},...,s_{iM}\}$ for the $i$-th variable where $M$ is the number of all detected objects in all the frames in the validation set. Given an error rate $\alpha$, we select the quantile $\hat{q}_i$ as the $\frac{\lceil (1-\alpha)(1+M) \rceil}{M}$ quantile of the score set. The prediction set for $x_i$ is constructed following the proposition.

\begin{prop}
When we assume the uncertain scalar for $i$-th variable follows the Gaussian distribution with mean $\hat{y}_i(x_i)$ and standard deviation $\hat{\sigma}_i(x_i)$ and select the score function as $s(x_i, y_i) = \frac{\left| y_i - \hat{y}_i(x_i) \right|}{\hat{\sigma_i}(x_i)}$ in CP, the prediction set for $i$-th variable is $\mathcal{C}_i(x_i) = \left[\hat{y}_i(x_i) - \hat{\sigma}_i(x_i)\hat{q}_i, \hat{y}_i(x_i) + \hat{\sigma}_i(x_i)\hat{q}_i \right]$.
\end{prop}
\begin{proof}
From Lemma~\ref{lemma:coverage}, for a test point $(X_{test}= x_i, Y_{test} = y_i)$, it holds that 
\begin{align}
    & \Pr(Y_{test} \in \mathcal{C}(X_{test}))  \geq 1 - \alpha \nonumber \\
\Rightarrow & \Pr( s(X_{test}, Y_{test}) \leq \hat{q}_i) \geq 1- \alpha  ~~~~(\text{Eq.~\eqref{equ:prediction_set}}) \nonumber \\
\Rightarrow & \Pr( \frac{|y_i - \hat{y}_i(x_i)|}{\hat{\sigma}_i(x_i)} \leq \hat{q}_i) \geq 1- \alpha \nonumber ~~~~(\text{Eq.~\eqref{eq:score_func}})\\
\Rightarrow & \Pr( |y_i - \hat{y}_i(x_i)| \leq \hat{\sigma}_i(x_i)\hat{q}_i  ) \geq 1- \alpha \nonumber\\
\Rightarrow & \mathcal{C}_i(x_i) = \left[\hat{y}_i(x_i) - \hat{\sigma}_i(x_i)\hat{q}_i, \hat{y}_i(x_i) + \hat{\sigma}_i(x_i)\hat{q}_i \right].
\end{align}
\end{proof}


Then we adjust the standard deviation by $\hat{\sigma}_i = \hat{\sigma}_i \hat{q}_i$ to achieve rigorously estimated uncertainty.


\subsection{Uncertainty Propagation to MOT}
\label{subsec:uq_track}
After obtaining the corrected standard deviation for each variable of detected objects, how to utilize and propagate it into the MOT stage remains a significant challenge. Here, we propose SDKF and NLLAI methods to leverage the uncertainty in both the motion prediction and association which are the primary steps of MOT.

\textbf{Standard Deviation-based Kalman Filter (SDKF):} As shown in Section~\ref{sec:relatedwork}, Kalman Filter (KF)~\cite{kalman1960new} is one important step for motion prediction. The inputs of KF encompass the observed state and measurement uncertainty. Compared to the existing MOT algorithm, we leverage our rectified standard deviation as the measurement uncertainty in place of the pre-established values. By taking into account both the mean and standard deviation of the detections, we are able to better account for the uncertainty of objects and provide more robust tracklets over time. SDKF does not significantly impact the time complexity of algorithms, as it only modifies the measurement uncertainty input from fixed values to rigorously estimated ones.

\vspace{-10pt}
\begin{algorithm}
\small
\caption{NLL-based Association Improvement method (NLLAI)}
\label{alg:nllai}
\KwData{Matched detection and tracklet list $\mathcal{A}_{matched}$, unmatched detection list $\mathcal{D}_{unmatched}$, unmatched tracklet list $\mathcal{T}_{unmatched}$, NLL threshold $\tau$}
\KwResult{New association results $\mathcal{A}'_{matched}$, $\mathcal{D}'_{unmatched}$, $\mathcal{T}'_{unmatched}$}

Similarity matrix $SNLL = NLL(\mathcal{D}_{unmatched}, \mathcal{T}_{unmatched})$ \label{alg:line_snll} \\
$\mathcal{A}'_{matched} \leftarrow$ assoicate $\mathcal{D}_{unmatched}$ and $\mathcal{T}_{unmatched}$ by Hungarian Algorithm with $SNLL$\label{alg:line_hungarian}\\
$\mathcal{D}'_{unmatched} \leftarrow \emptyset, \mathcal{T}'_{unmatched} \leftarrow \emptyset$\\
\For{ $(d, t)$ in $\mathcal{A}'_{matched}$}{\label{alg:line_unmatched_begin}
\If{$SNLL(d,t) > \tau$}{
    $\mathcal{A}'_{matched} = \mathcal{A}'_{matched} \backslash \{(d,t)\}$\\
    $\mathcal{D}'_{unmatched} = \mathcal{D}'_{unmatched} \cup \{d\}$\\
    $\mathcal{T}'_{unmatched} = \mathcal{T}'_{unmatched} \cup \{t\}$\\
}
}\label{alg:line_unmatched_end}

$\mathcal{A}'_{matched} = \mathcal{A}_{matched} \cup \mathcal{A}'_{matched}$\\

\end{algorithm}
\vspace{-10pt}

\textbf{Negative Log Likelihood-based Association Improvement (NLLAI):} Using Intersection over Union (IoU)-based similarity score cannot match low-quality detection results as shown in Section~\ref{sec:intro}, which poses a significant challenge during the association stage. To address this issue, we propose the NLLAI technique as shown in Algorithm~\ref{alg:nllai}. We first define Negative Log Likelihood (NLL) between the predicted locations of tracklets and the detected locations as a novel similarity score:
\vspace{-10pt}
\begin{equation}
    snll = - \frac{1}{I} \sum_{i=1}^I \log P(\dot{y}_i|\hat{y}_i, \hat{\sigma}_i),
\label{eq:nll_sim}
\end{equation}
where $\dot{y}_i$ is the predicted value for the $i$-th variable of the tracklet from the motion prediction model such as KF. As Subsection~\ref{subsec:uq_detection}, the distribution of each $i$-th variable for detected objects is a single-variate Gaussian distribution where $\hat{y}_i$ is the mean and $\hat{\sigma}_i$ is the standard deviation. Given the set of unmatched detections and unmatched tracklets after the original association method, we compute the NLL similarity matrix $SNLL$ with Equation~\ref{eq:nll_sim} (see Line~\ref{alg:line_snll}). 

Then we utilize the Hungarian algorithm to establish associations between unmatched detections and unmatched tracklets based on $SNLL$.
To eliminate matched pairs with high NLL scores, we introduce a hyperparameter denoted by $\tau$ as the NLL score threshold. Specifically, any matched pairs with $s_{NLL} > \tau$ shall be deemed ineligible for further consideration (see Lines~\ref{alg:line_unmatched_begin}-\ref{alg:line_unmatched_end}). 

The time complexity of NLLAI depends on the number of input unmatched detections $N_D$ and the number of input unmatched tracklets $N_T$. The time complexity of computing NLL can be optimized to be $O(1)$~\cite{numerical3}, so computing the similarity matrix needs $O(N_DN_T)$ time. Assuming $N_D>N_T$, the time complexity of associating with the Hungarian Algorithm can be $O(N_D^3)$~\cite{kuhn1955hungarian}. Thus the time complexity of our NLLAI is $O(N_D^3)$ which is polynomial.

\section{Experiment}
\label{sec:experiment}

\begin{figure}[t]
    \centering    
    \includegraphics[width=0.48\textwidth]{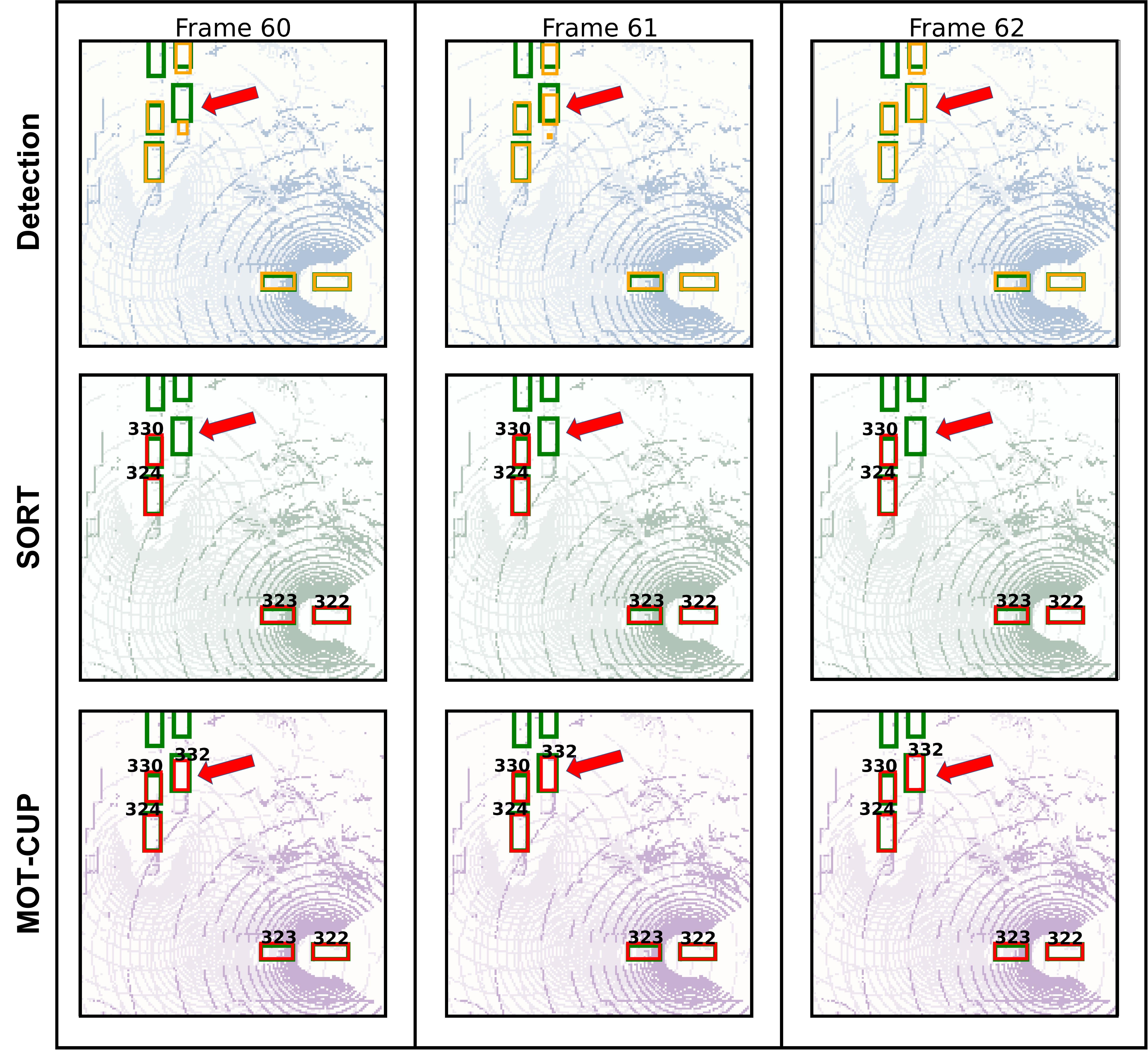}
    \caption{\textbf{Visualization of results of the detection, original SORT, and our MOT-CUP framework over consecutive three frames.} The collaborative object detector here is Upper-bound. Green boxes are ground truth bounding boxes, orange boxes are detected bounding boxes, and red boxes are tracklets' bounding boxes as the output of MOT. The numbers next to the red boxes indicate object IDs. We observe that our MOT-CUP outperforms the original SORT algorithm in tracking object 332, as indicated by the red arrow. Furthermore, MOT-CUP improves the accuracy of location, compared with the object detector, such as object 332 in frame 60. Overall, our results demonstrate the importance of considering uncertainty in MOT.}
    \label{fig:exp}
    \vspace{-20pt}
\end{figure}

\subsection{Experimental Setups}
We evaluate the uncertainty propagation framework MOT-CUP using the V2X-Sim dataset~\cite{li2022v2x}, which comprises 80 scenes for training, 10 scenes for validation, and 10 scenes for testing. V2X-Sim was generated using the CARLA simulation~\cite{Dosovitskiy17}. Each scene includes 
100 time-series frames and features 2-5 connected vehicles, from which 3D point clouds are collected using LiDAR sensors. Except for V2X-Sim, there are currently no other open-source datasets tailored explicitly to support COD and MOT. Therefore, our experiments focus solely on utilizing the V2X-Sim dataset. The host machine is a server with Intel Core i9-10900X processors and four NVIDIA Quadro RTX 6000 GPUs.

MOT methods use the tracking-by-detection framework. Object detection quality has a significant impact on tracking performance~\cite{bewley2016simple}. We consider three collaborative object detectors for all tracking approaches as follows:

   \textit{ Lower-bound (LB)}~\cite{li2022v2x}: The single-agent object detector, which operates independently by utilizing point cloud data from one single LiDAR sensor without the need for collaboration with other detectors.
    
    \textit{DiscoNet (DN)}~\cite{li2021learning}: The intermediate collaborative object detector employs a directed graph with matrix-valued edge weight to dynamically aggregate features from various agents.
    It demonstrates a favorable trade-off between performance and bandwidth. 
    
    \textit{Upper-bound (UB)}~\cite{li2022v2x}: The early collaborative object detector employs raw point cloud data from all connected vehicles to facilitate collaboration. This detector achieves high performance, while retaining lossless information. However, the approach often requires high communication bandwidth.

We apply our uncertainty propagation framework to two tracking baselines, SORT~\cite{bewley2016simple} and ByteTrack~\cite{zhang2021bytetrack}, and compare their performance in accuracy and uncertainty.
SORT~\cite{bewley2016simple} is a pragmatic approach with simple, effective algorithms by using the Kalman Filter for estimation and the Hungarian algorithm for data association. Instead of only associating detection boxes with high scores, ByteTrack~\cite{zhang2021bytetrack} also utilizes similarities between the low score detection boxes and tracklets 
to improve the performance on data association. In our MOT-CUP, we select the NLL threshold  $\tau=1000$ for SORT and $\tau=80$ for ByteTrack. Other hyperparameters such as the IoU threshold, are directly inherited from the original designs of \cite{li2022v2x, zhang2021bytetrack, bewley2016simple, angelopoulos2021gentle}.

\subsection{Evaluation Metrics}

\noindent \textbf{Accuracy Metrics}:  

    \textit{Higher Order Tracking Accuracy (HOTA)}~\cite{luiten2021hota}: captures the effect of accurate object detection, association, and localization in a well-balanced way. Such a unified measure captures the synergistic impact of these critical aspects and most comprehensively assesses the algorithm's effectiveness.
    
    \textit{Multiple Object Tracking Accuracy (MOTA)}~\cite{bernardin2008evaluating}: quantifies missed detections, false positives and false negatives for detection, and identity switches for the association.
    
    \textit{Multiple Object Tracking Precision (MOTP)}~\cite{bernardin2008evaluating}: measures the ability to estimate precise object locations.
    
    \textit{Frames Per Second (FPS)}: refers to the number of frames processed per second, and measures the time complexity.

It is important to note that higher values of the aforementioned performance metrics indicate better performance in the context of MOT evaluation. 
When assessing the performance of MOT algorithms on the same object detector, even slight improvements in HOTA, MOTA, and MOTP mean good progress, as reported by~\cite{zhang2021bytetrack, lakshminarayanan2017simple}.

\noindent \textbf{Uncertainty Metrics}:

    \textit{Negative Log Likelihood (NLL)}~\cite{harakeh2021estimating}: a prevalent metric employed to assess the level of uncertainty in the predicted probability distribution of a given test dataset~\cite{boris2022propagating, lakshminarayanan2017simple, meyer2020learning}.
    
    \textit{Continuous Ranked Probability Score (CRPS)}~\cite{v2019online}: measures the discrepancy between predicted and ground-truth probability distributions~\cite{chung2021uncertainty,korotin20a}. 

 
Lower values indicate more precise uncertainty estimation.

\subsection{Accuracy Evaluation}\label{subsec:acc}

\begin{table}
\caption{Performance Evaluation of our Uncertainty Propagation Framework on different MOT baselines and object detectors}
\vspace{-15pt}
\label{tab:accuracy}
\tabcolsep=2pt
\begin{center}
\begin{tabular}{|c|c|c|c|c|c|c|}
\hline
Base & Detector & Method & HOTA $\uparrow$& MOTA $\uparrow$& MOTP $\uparrow$& FPS$\uparrow$\\
\hline
 & \multirow{2}{*}{UB} & Base & 41.34 & 52.60 & 85.42 & 1026\\
\cline{3-7}
& & Our & 42.19 & 53.72 & 86.07& 877 \\
\cline{2-7}
SORT& \multirow{2}{*}{DN} & Base & 41.80 & 50.79 & 85.42 & 1052 \\
\cline{3-7}
\cite{bewley2016simple}& & Our & 42.49 & 51.73 & 85.85 & 885 \\
\cline{2-7}
& \multirow{2}{*}{LB} & Base & 31.28 & 27.09 & 85.62 & 1568 \\
\cline{3-7}
& & Our & 31.69 & 27.70 & 85.69 & 1317 \\
\hline
 & \multirow{2}{*}{UB} & Base & 42.05 & 52.90 & 84.47 & 1251 \\
\cline{3-7}
& & Our & 42.56 & 53.77 & 85.49 & 1067 \\
\cline{2-7}
Byte-& \multirow{2}{*}{DN} & Base & 42.64 & 51.48 & 84.01 & 1153 \\
\cline{3-7}
Track& & Our & 43.14 & 52.28 & 84.94 & 1074 \\
\cline{2-7}
\cite{zhang2021bytetrack}& \multirow{2}{*}{LB} & Base & 32.27 & 29.15 & 84.57 & 1637 \\
\cline{3-7}
& & Our & 32.55 & 29.65 & 84.97 & 1457 \\
\hline
\end{tabular}
\vspace{-15pt}
\end{center}
\end{table}

\begin{table}[ht]
\caption{Performance evaluation of our MOT-CUP Framework on various occlusion-level scenarios.}
\vspace{-15pt}
\label{tab:occlusion}
\tabcolsep=2pt
\begin{center}
\begin{tabular}{|c|c|c|c|c|}
\hline
Scenario & Method & HOTA $\uparrow$& MOTA $\uparrow$& MOTP $\uparrow$\\
\hline
High & Base & 29.00 & 34.57 & 68.13  \\
\cline{2-5}
OCL & Our & 30.16 (4.01\%) & 36.14 (4.52\%) & 68.85 (1.06\%) \\
\hline
Low & Base & 43.51 & 57.24 & 85.51  \\
\cline{2-5}
OCL & Our & 44.25 (1.71\%) & 58.03 (1.39\%) & 86.07 (0.67\%) \\
\hline
\end{tabular}
\vspace{-25pt}
\end{center}
\end{table}

\begin{table*}
  \centering
  \caption{NLL \& CRPS comparisons on detection and MOT-CUP with different uncertainty quantification methods: Dropout (DO), Deep Ensemble (DE) and Conformal Prediction (CP). The best results are shown in \textbf{bold}.}
  \tabcolsep=3pt
  \vspace{-10pt}
  \label{tab:nll}
  \begin{tabular}{|c|c|c|c|c|c|c|c|c|c|c|c|c|c|c|c|c|}
  \hline
   \multirow{2}{*}{Base} & \multirow{2}{*}{Method} & \multirow{2}{*}{DO} & \multirow{2}{*}{DE} & \multirow{2}{*}{CP}& \multicolumn{3}{c|}{NLL @IoU=0.5 $\downarrow$}&\multicolumn{3}{c|}{NLL @IoU=0.7 $\downarrow$} & \multicolumn{3}{c|}{CRPS @IoU=0.5 $\downarrow$}&\multicolumn{3}{c|}{CRPS @IoU=0.7 $\downarrow$}\\
  \cline{6-17}
  & & &&&UB& DN & LB& UB& DN& LB & UB& DN & LB& UB& DN& LB \\
  \hline
  & \multirow{4}{*}{Detection}& &&&193 & 222 & 335 & 96 & 147 & 166 & 0.453 & 0.498 & 0.693 & 0.392 & 0.459 & 0.514\\
  \cline{3-17}
  &&\checkmark&&&81.01 & 52.97 & 263 & 36.05 & 26.98 & 124 & 0.512 & 0.554 & 0.745 & 0.453 & 0.517 & 0.572\\
  \cline{3-17}
  &&&\checkmark&&44.01 & 46.98 & 128 & 23.64 & 29.30 & 63.80 & 0.482 & 0.518 & 0.703 & 0.423 & 0.480 & 0.531 \\
  \cline{3-17}
  SORT& & & & \checkmark & 25.70 & 26.21 & 25.17 & 14.54 & 19.50 & 13.04 & 0.424 & 0.466 & 0.652 & 0.364 & 0.427 & 0.475\\
  \cline{2-17}
  \cite{bewley2016simple}&\multirow{4}{*}{MOT-CUP}  &&&  & 9.61 & 12.09 & 14.45 & 7.87 & 11.21 & 11.06 & 0.312 & 0.355 & 0.463 & 0.297 & 0.347 & 0.409\\
  \cline{3-17}
  &&\checkmark&&&6.54 & 3.48 & 18.34 & 4.38 & 3.44 & 11.40 & 0.345 & 0.379 & 0.489 & 0.331 & 0.374 & 0.436 \\
  \cline{3-17}
  &&&\checkmark&&2.56 & 2.31 & 6.15 & 2.17 & 2.23 & 3.82 & 0.338 & 0.360 & 0.483 & 0.324 & 0.354 & 0.431 \\
  \cline{3-17}
  &  &&& \checkmark& \textbf{0.94} & \textbf{0.95} & \textbf{1.30} & \textbf{0.74} & \textbf{0.90} & \textbf{1.06} & \textbf{0.301} & \textbf{0.336} & \textbf{0.444} & \textbf{0.286} & \textbf{0.329} & \textbf{0.392} \\
  \hline
  & \multirow{4}{*}{Detection} &&&& 1801 & 540 & 461 & 870 & 198 & 121 & 0.392 & 0.391 & 0.597 & 0.338 & 0.357 & 0.358 \\
  \cline{3-17}
  &&\checkmark&&&101& 35.08 & 269 & 41.17 & 24.86 & 101 & 0.524 & 0.561 & 0.792 & 0.468 & 0.531 & 0.550\\
  \cline{3-17}
  &&&\checkmark&&72.56 & 38.41 & 156.59 & 35.99 & 30.51 & 57.10 & 0.492 & 0.523 & 0.753 & 0.436 & 0.493 & 0.518\\
  \cline{3-17}
  Byte-& & &&\checkmark& 43.65 & 32.50 & 68.94 & 23.47 & 19.25 & 11.78 & 0.381 & 0.376 & 0.596 & 0.328 & 0.343 & 0.358 \\
  \cline{2-17}
  Track& \multirow{4}{*}{MOT-CUP}&&& & 29.49 & 25.30 & 57.23 & 17.67 & 17.57 & 11.27 & 0.302 & 0.318 & 0.412 & 0.276 & 0.308 & 0.310 \\
  \cline{3-17}
  \cite{zhang2021bytetrack}&&\checkmark&&&24.35 & 7.66 & 26.43 & 6.97 & 7.37 & 21.70 & 0.385 & 0.436 & 0.532 & 0.359 & 0.432 & 0.437 \\
  \cline{3-17}
  &&&\checkmark&&23.52 & 11.61 & 24.62 & 6.20 & 9.89 & 20.77& 0.362 & 0.406 & 0.512 & 0.336 & 0.402 & 0.421 \\
  \cline{3-17}
  & &&&\checkmark& \textbf{20.01} & \textbf{6.94} & \textbf{4.05} & \textbf{2.41} & \textbf{1.99} & \textbf{0.99} & \textbf{0.280} & \textbf{0.286} & \textbf{0.376} & \textbf{0.254} & \textbf{0.275} & \textbf{0.276}\\
  \hline
  \end{tabular}
  \vspace{-5pt}
\end{table*}

\begin{table*}
\caption{Ablation Study on MOT with the Upper-bound and DiscoNet detectors. The best results are shown in \textbf{bold}.}
\label{tab:ablation_ub}
\vspace{-15pt}
\begin{center}
\begin{tabular}{|c|c|c|c|c|c|c|c|c|c|c|c|}
\hline
\multirow{2}{*}{Base} & \multirow{2}{*}{CP} & \multirow{2}{*}{SDKF} & \multirow{2}{*}{NLLAI} & \multicolumn{4}{c|}{Upper-bound} & \multicolumn{4}{c|}{DiscoNet} \\
\cline{5-12}
 &  &  &  & HOTA $\uparrow$& MOTA $\uparrow$& MOTP $\uparrow$& FPS $\uparrow$& HOTA $\uparrow$& MOTA $\uparrow$& MOTP $\uparrow$& FPS $\uparrow$\\
\hline
& & & & 41.34 & 52.60 & 85.42 & \textbf{1026} & 41.80 & 50.79 & 85.42 & \textbf{1052}\\
\cline{2-12}
& & \checkmark & & 41.67 & 52.35 & 86.25 & 962 & 42.23 & 50.79 & 86.15 & 985\\
\cline{2-12}
& & & \checkmark & 41.34 & 52.60 & 85.42 & 865 & 41.80 & 50.79 & 85.42 & 873 \\
\cline{2-12}
SORT& \checkmark& \checkmark& & 41.80 & 52.65 & 86.20 & 991 & 42.23 & 50.93 & 86.09 & 993 \\
\cline{2-12}
\cite{bewley2016simple}& \checkmark& &\checkmark & 41.73 & 53.50 & 84.81 & 911 & 41.98 & 51.32 & 84.70 & 899 \\
\cline{2-12}
& & \checkmark& \checkmark& 41.67 & 52.35 & \textbf{86.25} & 841 & 42.23 & 50.80 & \textbf{86.15} & 852 \\
\cline{2-12}
& \checkmark& \checkmark& \checkmark& \textbf{42.19} & \textbf{53.72} & 86.07 & 877 & \textbf{42.49}& \textbf{51.73} & 85.85 & 885 \\
\hline
& & & & 42.05 & 52.90 & 84.47 & \textbf{1251} & 42.64 & 51.48 & 84.01 & \textbf{1153} \\
\cline{2-12}
& & \checkmark & & 42.49 & 53.63 & 85.82 & 1195 & 42.85 & 52.13 & 85.29 & 1078 \\
\cline{2-12}
Byte-& & & \checkmark & 42.06 & 52.97 & 84.45 & 1072 & 42.67 & 51.61 & 83.98 & 1010 \\
\cline{2-12}
Track& \checkmark& \checkmark& & 42.62 & 53.85 & 85.54 & 1219 & 43.09 & 52.21 & 85.04 & 1141 \\
\cline{2-12}
\cite{zhang2021bytetrack}& \checkmark& &\checkmark & 42.12 & 53.21 & 84.30 & 1057 & 42.68 & 51.65 & 83.89 & 1029 \\
\cline{2-12}
& & \checkmark& \checkmark& 42.50 & 53.68 & \textbf{85.80} & 1091 & 42.89 & 52.26 & \textbf{85.26} & 1042 \\
\cline{2-12}
& \checkmark& \checkmark& \checkmark& \textbf{42.56} & \textbf{53.77} & 85.49 & 1067 & \textbf{43.14} & \textbf{52.28} & 84.94 & 1074\\
\hline
\end{tabular}
\vspace{-25pt}
\end{center}
\end{table*}

The outcomes of our framework on the V2X-SIM dataset with three distinct object detectors and two diverse MOT baselines are presented in Table~\ref{tab:accuracy}. The results indicate that our MOT-CUP framework is capable of leveraging quantified uncertainty from COD to enhance the performance of all original MOT algorithms, with up to $0.85$ improvement on HOTA, up to $1.13$ improvement on MOTA and up to $1.03$ improvement on MOTP.  The performance of object detectors can significantly impact the performance of  MOT. Specifically, when the object detector is capable of detecting more objects, such as Upper-bound, our framework can significantly enhance the performance of MOT algorithms. 

\textit{MOT-CUP on high occlusion-level scenarios:} We divide the entire test dataset into two subsets: one with high occlusion scenarios and the other with low occlusion scenarios. We conduct experiments using our MOT-CUP with SORT and Upper-bound, as in Table~\ref{tab:occlusion}. The results demonstrate that our MOT-CUP exhibits superior improvements in high occlusion-level scenarios, with a notable 4.01\% enhancement in HOTA compared to a 1.71\% improvement in HOTA for low occlusion-level scenarios. In high occlusion-level scenarios, the presence of poorly detected objects caused by occlusion 
leads to high uncertainty, which our SDKF and NLLAI 
utilize to enhance the tracking performance.

Fig.~\ref{fig:exp} presents visualizations of Upper-bound, original SORT, and our MOT-CUP framework's results over three consecutive frames. Our MOT-CUP outperforms the original SORT in tracking object 332, as indicated by the red arrow. Moreover, our MOT-CUP improves location accuracy, as shown for object 332 in frame 60, compared to the object detector. These results showcase the effectiveness of our approach in accurately tracking objects, even in challenging scenarios with poor detection or occlusion. Additionally, incorporating uncertainty into the Kalman Filter and association step enables better tracking performance over time.

\subsection{Uncertainty Evaluation}\label{subsec:uncertainty}

We use Negative Log Likelihood (NLL)~\cite{harakeh2021estimating} and Continuous Ranked Probability Score (CRPS)~\cite{v2019online} at IoU thresholds of 0.5 and 0.7 as the uncertainty measurement. 
We compare the uncertainty results of different UQ methods on object detection and MOT-CUP, including dropout (DO), deep ensemble (DE), and our conformal prediction (CP) in Table~\ref{tab:nll}. The vanilla baseline only utilizes direct modeling (DM). The implementations of DO and DE are as same as~\cite{feng2021review}.
The representation formats of bounding boxes utilized by SORT and ByteTrack diverge, necessitating the training of distinct detection models. Consequently, the results on uncertainty are dissimilar between SORT and ByteTrack.


For NLL, CP outperforms all baselines, with up to $41\times$ improvement. In particular, CP achieves up to 95\% reduction compared to DO and DE. Furthermore, in comparison to object detection, our MOT-CUP framework with CP produces precise uncertainty estimation, with up to $2.67 \times$ improvement. The vanilla object detection shows a significantly large NLL for the DM overfits the training dataset and overestimates the uncertainty of the test dataset.


Compared with all baselines, CP can effectively reduce the CRPS, with up to 37\% reduction. Specially, it achieves up to 31\%, 37\% and 35\% reduction compared with the vanilla baseline, DO and DE. The reason that DO and DE increase the CRPS might be they cannot fully capture the entire distribution of possible values while CRPS requires the entire predicted distribution to be considered~\cite{chung2021uncertainty}.

\subsection{Ablation Study on Accuracy}\label{subsec:ablation}

We conduct an ablation study to evaluate the contributions of each proposed technique in our MOT-CUP framework as shown in Table~\ref{tab:ablation_ub} with two different detectors and two diverse MOT baselines. CP is shown to contribute significantly to both SDKF and NLLAI. NLLAI, which focuses on refining the association step with a new SNLL metric, yields marked improvements in metrics capturing associations such as HOTA and MOTA. However, an increase in matching potential may lead to a decline in the precision of object localization, as reflected by the decrease in MOTP metric. In contrast, SDKF, where the Kalman Filter takes the COD uncertainty information as its input,  primarily enhances metrics measuring localization, such as HOTA and MOTP, thereby improving the accuracy of object localization estimates. Notably, our proposed framework combined with diverse collaborative object detectors and MOT baselines always achieves the optimal performance outcomes.

\paragraph{Limitation}
In terms of FPS, our framework results in an average decrease of 13.2\%, yet it does not affect the real-time capacity of the MOT algorithms.  It is noteworthy that the increase in time incurred by our framework is polynomial, as discussed in Section~\ref{sec:approach}. Furthermore, we have not implemented any specific strategies aimed at optimizing the quality of code with respect to running time. Therefore, the computational overhead of our framework is acceptable.

\section{Conclusion}
\label{sec:conclusion}
This paper presents the first attempt to leverage uncertainty quantification from collaborative object detection (COD) to enhance the performance of multiple object tracking (MOT). Our proposed framework, MOT-CUP, employs direct modeling and conformal prediction techniques to quantify the uncertainty in COD. The uncertainty of COD is propagated to the Kalman Filter and the Negative Log Likelihood-based Association Improvement (NLLAI) procedure of MOT. We evaluate MOT-CUP on various CODs and MOT baselines, and demonstrate that our framework significantly improves both the accuracy and uncertainty of the original MOT. Our findings highlight and validate the benefits of incorporating COD uncertainty quantification into MOT algorithms. In future work, we plan to extend our method to popular single-agent object detection and MOT benchmarks, such as KITTI and nuScenes, and more MOT baselines.


\bibliographystyle{IEEEtran}
\bibliography{uncertainty}


\end{document}